\documentclass[preprint]{elsart}

\usepackage{epsfig}

\usepackage{amssymb}
\usepackage{amsmath}

\begin{document}

\begin{frontmatter}


\title{A Hybrid Artificial Immune System and Self Organising Map for Network Intrusion Detection}
\author{Simon T. Powers\corauthref{cor1}}
\corauth[cor1]{Corresponding author.}
\address{School of Electronics and Computer Science, University of Southampton, Highfield, Southampton, SO17 1BJ, U.K.}
\ead{stp@soton.ac.uk}


\author{Jun He}
\address{Department of Computer Science, University of Wales, Aberystwyth, SY23 3DB, U.K.}
\ead{jqh@aber.ac.uk}

\begin{abstract}
Network intrusion detection is the problem of detecting unauthorised use of, or access to, computer systems over a network. Two broad approaches exist to tackle this problem: anomaly detection and misuse detection. An anomaly detection system is trained only on examples of normal connections, and thus has the potential to detect novel attacks. However, many anomaly detection systems simply report the anomalous activity, rather than analysing it further in order to report higher-level information that is of more use to a security officer. On the other hand, misuse detection systems recognise known attack patterns, thereby allowing them to provide more detailed information about an intrusion. However, such systems cannot detect novel attacks. 

A hybrid system is presented in this paper with the aim of combining the advantages of both approaches. Specifically, anomalous network connections are initially detected using an artificial immune system. Connections that are flagged as anomalous are then categorised using a Kohonen Self Organising Map, allowing higher-level information, in the form of cluster membership, to be extracted. Experimental results on the KDD 1999 Cup dataset show a low false positive rate and a detection and classification rate for Denial-of-Service and User-to-Root attacks that is higher than those in a sample of other works.     

[This is a post-print of an accepted manuscript published in \textit{Information Sciences} 178(15), pp. 3024–-3042, August 2008. The publisher's version is available at http://www.sciencedirect.com/science/article/pii/S0020025507005531.]

\end{abstract}

\begin{keyword}
Artificial immune system; Self Organizing Map; Intrusion detection; Genetic algorithm; Negative Selection; Anomaly detection
%\PACS 
\end{keyword}
\end{frontmatter}

% main text
\section{Introduction}
\label{secIntro}

Ensuring the accessibility, confidentiality, and integrity of data stored on computer systems is an ever growing cause of concern, given the opportunities for malicious activity afforded by global connectivity. One approach would be to attempt to build completely secure systems, i.e. software systems without vulnerabilities; however, this is unlikely to come to fruition given the inherent difficulty in building such systems and the costs involved \cite{Denning:1987:a,Mukherjee:1994:a}. It is therefore inevitable for the foreseeable future that breaches of system security will occur. The pertinent question is then how to detect such breaches so that appropriate actions, such as removing the intruder and reporting him to the appropriate authorities, can be performed. 

An intrusion detection system (IDS) is designed to detect unauthorised use of, or access to, a computer system by both those with legitimate user accounts and those from outside with no access rights \cite{Mukherjee:1994:a}. Two broad categories of IDS exist; anomaly detection systems build a model of normal system activity and then regard deviations from this as potential intrusions \cite{Denning:1987:a}, while misuse detection systems look for known attack patterns by signature matching. The key advantage of anomaly detection systems over signature based misuse approaches is their ability to detect novel attack patterns for which no signature exists, while their most often cited disadvantage is a larger false positive rate. However, in this paper we aim to address a second, often overlooked, disadvantage of a pure anomaly detection system. This is that such systems usually simply report the anomalous actions; they do not provide any further information as to the consequences of the attack and the possible future actions of the attacker.     

In order to achieve this, a novel hybrid IDS has been developed that uses two different nature-inspired techniques. In the first stage, an artificial immune system (AIS) is used to identify anomalous TCP/UDP connections made to machines on a network. High-level information about identified anomalous connections is then provided by clustering the anomaly into one of a number of broad attack types, using a Kohonen Self Organising Map (SOM) \cite{Kohonen:1982:a}. 

The following sections describe in further detail the motivation for the hybrid approach used and the contributions of this paper.

\subsection{Motivation for a two-stage approach to intrusion detection}

The motivation for a two-stage approach to intrusion detection comes from considering the advantages and disadvantages of a pure anomaly detection system. An anomaly detection system works by constructing a model of normal activity, for example, a model of normal TCP/UDP connections to a machine. New activity that deviates from this model is then considered to potentially signify an intrusion. Clearly, such systems do not depend upon any a priori knowledge of possible attack patterns, since they are constructed solely from examples of normal activity. Consequently, they can detect novel attack methods, providing that the actions carried out in performing the attack deviate sufficiently from the model of normal. 

This potential to detect novel attack methods is highly desirable, given the large rate at which they are discovered and utilised. For example, the CERT team at Carnegie Mellon University reported 5990 previously unknown security vulnerabilities in 2005 alone \cite{Cert:2006:a}, any of which could potentially be exploited by an attacker. This large rate of occurrence of novel attacks suggests that the purely reactive approach of comparing current activity against known patterns of misuse (e.g. as used by the popular open source IDS SNORT \cite{Snort:2006:a}) is not sufficient to protect systems from malicious attackers. Therefore, anomaly detection systems that are scalable up to real-world implementations seem to be a promising avenue of research. 

However, there are two reasons why stand-alone anomaly detection systems are not often used in a real-world IDS, the first of which has already been widely discussed in the literature. The first problem is the inherent higher false positive rate compared to a system such as SNORT that simply looks for the signatures of known attack patterns. A false positive is where the IDS flags normal activity as anomalous. Such incidents waste the time of human security officers and can cause confidence in the system to be lost, potentially leading to real attacks that are flagged by the system being ignored. It is therefore very important for an IDS to have a low false positive rate. The reason for the higher false positive rate of anomaly detection systems is that users sometimes perform new and different activities, making it very hard to build a model of normal that is broad enough to encompass such actions but not so broad as to also mistakenly count attack patterns as normal. The fundamental problem is that regardless of the particular activity level monitored by the IDS, be it network packets or operating system audit trails, the root cause of that activity is a human, and humans are notoriously unpredictable. 

The second reason, and the motivation for the inclusion of a second component, has to our knowledge not explicitly been commented on in the literature. This is that all a stand-alone pure anomaly detection system can do when an anomaly occurs is to flag the anomalous actions for the attention of the human security officer; such a system cannot alone provide any further high-level information. The drawback with only providing low-level information such as a list of anomalous actions is that such information is very tedious and time consuming for a security officer to process. This problem of overwhelming the security officer with low-level alerts has been recognised in recent years in the literature on data fusion \cite{Bass:1999:a,Bass:2000:a}, which proposes combining multiple alerts from different sensors (components of an IDS) into more meaningful high-level alerts. The difference between the approach proposed in this paper and data fusion techniques is that we consider how to derive high-level information from a \emph{single} low-level sensor, namely one particular anomaly detection system. Of course, such derived higher-level information from a single sensor could then be combined with the output of other sensors, such as another anomaly detection system operating on OS audit trails, using a data fusion approach.

\subsection{Contributions of this paper}     

The remaining question is then how to derive higher-level information from reported anomalous actions. The idea used in this paper is to analyse examples of known attacks for common statistical patterns or features, thereby allowing clusters of attacks sharing similar properties to be created. A cluster centre then contains common properties of many attacks, i.e. it represents a higher-level abstraction of those attacks. After the cluster centres have been created, an anomalous activity can be matched to the cluster that it is most similar to. Cluster matching means that the anomaly shares common properties with other attacks that belong to that cluster. Therefore, rather than simply reporting the anomaly, the IDS can provide abstract high-level information in the form of properties of the cluster centre.

Other works have also proposed the clustering of known attack patterns. For example, a SOM was used in \cite{DeLooze:2004:a} to cluster attacks from the CERT database in a similar manner. However, such an approach has not been used to complement a stand-alone anomaly detection system. Instead, the same clustering technique is usually used to not only cluster attack patterns but also to cluster normal activity, e.g. \cite{DeLooze:2006:a}. Likewise, multi-layer perceptron neural networks have been widely used to make the distinction between both normal and anomalous and then within anomalous between different attack classes (e.g. \cite{Mukkamala:2002:a,Li:2004:a}). 

Using such a single monolithic classifier that is trained on examples of both normal activity and attack patterns is not necessarily wise, since it may lead to biasing the classifier towards only recognising the example attack patterns that are present in the training data. As a result, the classifier may fail to detect novel attacks, thereby loosing the important advantage of anomaly detection systems. In other words, such a system will cease to perform anomaly detection but instead will perform misuse detection with a generalisation capability, as was the case in \cite{Cannady:1998:a}.

The approach developed in this paper should also be contrasted with systems that combine an anomaly detection component with a misuse component that is designed to recognise the signatures of known attacks, e.g. \cite{Depren:2005:a}. In such systems, the misuse component is used to ensure that known attacks are recognised, while the anomaly detection component is used to attempt to detect novel attack patterns. While this approach can help to improve the overall detection rate, it still suffers from the drawback that only low-level information is provided about novel attack patterns identified by the anomaly detection component.

\subsection{Overview of our novel hybrid IDS}

This section provides a brief overview of the runtime operation and training process of our hybrid IDS. 

At runtime, the system monitors incoming connections as follows. Firstly, a connection vector is created to represent an incoming connection. This consists of features describing the connection, such as the port number and number of packets sent. The connection vector then undergoes anomaly detection by detectors generated through negative selection. 

\begin{figure}[htbf]
	\epsfig{file=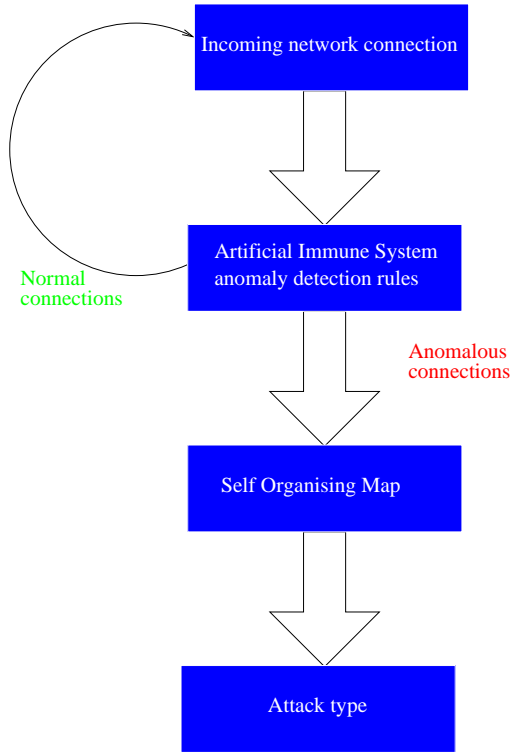,height=10cm}
% papRoc.eps: 300dpi, width=9.13cm, height=4.93cm, bb= -233   104   845   686
	\caption{Overview of the runtime operation of the hybrid IDS.}
	\label{figRuntime}
\end{figure}

Any connection vectors flagged as anomalous are then projected onto a SOM. This projection places the connection vector onto a neuron close to those that related attacks are projected onto. This therefore highlights the attacks that the connection shares common properties with. In this way, attacks are detected by the anomaly detectors, before being projected onto the SOM to determine which other attacks a new attack is most similar to, and hence provide higher-level information. This real-time operation of the system is highlighted in Figure ~\ref{figRuntime}.

During training, the two components of the system are trained independently. The anomaly detectors are produced by the negative selection algorithm, using only examples of \emph{normal} connections in their training data. Conversely, the SOM is trained exclusively on examples of \emph{anomalous} connections, i.e. attacks. This is because the SOM is charged with the task of clustering similar attacks together, extracting common features from related attacks. Note that unlike most other types of neural network, the training of the SOM is unsupervised. This means that the attack connections are not labelled with names or types during the computation of the neurons' weight vectors. The SOM therefore groups together attacks that share common properties with no a priori knowledge.

\begin{figure}[htbf]
	\epsfig{file=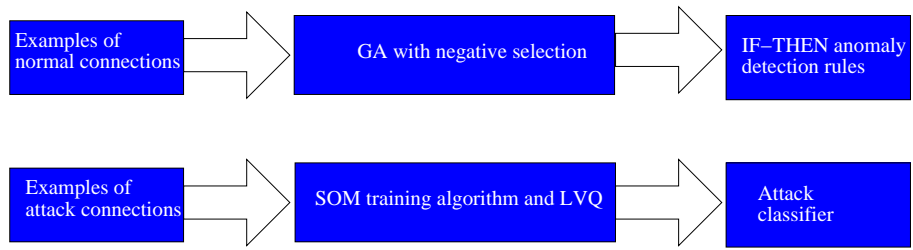,width=12cm}
% papRoc.eps: 300dpi, width=9.13cm, height=4.93cm, bb= -233   104   845   686
	\caption{Overview of the training procedure for the two components.}
	\label{figTrain}
\end{figure}

Figure~\ref{figTrain} illustrates the training procedure for the two components. The following sections describe the training processes in detail.

\section{Anomaly Detection Using an Artificial Immune System}
\label{secAnomalyDetection}

In recent years, the field of artificial immune systems has begun to flourish \cite{deCastro:2002:a}. An artificial immune system uses ideas from the operation of the human immune system and applies them to computational problems. Of particular relevance to the problem of intrusion detection is the fact that that the immune system can be viewed as performing anomaly detection, since it distinguishes between normal self and harmful non-self, e.g. pathogens or tumour cells, in the body. This anomaly detection is performed by a certain type of lymphocyte known as a T-cell, which circulates around the body.

Each T-cell contains a binding site that allows it to bind to certain antigens for which its binding site ``matches'' the antigen. Because one T-cell can only match certain antigens, a whole population of T-cells is required to protect the body from different non-self cells. The binding site of a T-cell, and hence which antigens can be matched by it, is determined during T-cell creation by a random genetic rearrangement process. A consequence of this random generation process is that a T-cell could potentially bind to self cells, i.e. raise a false alarm by detecting normal as anomalous. This is prevented by a maturation process, whereby any newly created T-cells that bind to self are destroyed, before they are allowed into the blood stream. This process of selecting only those T-cells that do not match self is termed negative selection. 

\subsection{Background} 

The research group of Forrest \cite{Forrest:1994:a} was the first to apply the idea of negative selection in a computational setting. In their algorithm, antigens, such as properties of a TCP connection, were represented as fixed length binary strings. Detectors, inspired by T-cells, were then also represented as binary strings of the same length. A detector string was said to match an antigen string if the two strings shared the same characters in an uninterrupted stretch of $r$ bits; this is known as the $r$-contiguous bits matching rule. Negative selection was used to generate the detectors by randomly generating strings and then discarding those that matched any of the normal antigens in the training data. Thereafter, the detectors that passed the negative selection filter were used to monitor new antigens; if a detector matched any antigen then that antigen was flagged as non-self, i.e. anomalous.

Amongst other applications, e.g. \cite{Forrest:1996:a}, Forrest's group have applied their algorithm to the problem of network intrusion anomaly detection. Their LISYS system \cite{Hofmeyr:1999:a,Balthrop:2002:a} encodes the source IP address, destination IP address and server-side port of TCP connections in a 49-bit binary string. A set of self strings is obtained by observing normal TCP connections over a period of time, while negative selection is used to generate detector strings that aim to match anomalous connections that may occur in the future.

Unfortunately, there are two problems with such an approach. Firstly, the use of binary strings and an $r$-contiguous bits matching rule makes it difficult to extract high-level domain knowledge from the detectors \cite{Gonzalez:2003:b}. For example, in an intrusion detection system, it is useful to be able to analyse the detectors that were activated during an attack, in order to discover properties of the attack. However, analysing the part of a binary string that matched part of another such string is unlikely to yield much useful domain knowledge. This is because both the representation and the matching rule are too low-level to facilitate such a process.

The second problem with LISYS is one of applicability to a real-world scenario. It is certainly the case that simply looking at the IP addresses and ports of a connection is insufficient to detect many types of attacks. However, adding further information about the connection to the detector and antigen strings would rapidly increase their length, given that binary coding is used. Furthermore, as the detectors come to store more information, it becomes questionable whether random detector generation would be feasible. For example, random detector generation was shown to be infeasible in \cite{Kim:2001:b} when 33 features of a network connection were used. This was due to the length of time required to find a detector that did not match self.

To overcome these problems, the use of real-valued anomaly detectors has been proposed \cite{Gonzalez:2002:c,Gonzalez:2002:j}. The most significant improvement that this approach offers over binary string representations is a distinction between a detector \textit{genotype} and a detector \textit{phenotype}. At the genotypic level, the detectors are vectors of real numbers. At the phenotypic level, they are interpreted as specifying intervals on the space of real numbers. These intervals are then read as conditions for an IF-THEN rule, where the consequent is that an anomaly has been detected. This means that an antigen vector is matched by a detector if the components of the antigen vector lie within the corresponding intervals specified by the detector. A genetic algorithm is then used to generate detectors that do not match self, rather than the random generation of Forrest.

In applying this idea to network intrusion detection, 3 features were used, corresponding to intervals on aggregate network traffic statistics \cite{Gonzalez:2002:c,Gonzalez:2002:j}. Specifically, the features used were the total number of packets, the number of ICMP packets, and the number of bytes of data transmitted, over a period of 1 second. In addition, a sliding time window was used to attempt to detect temporal anomalies, e.g. if the window size was 3 then the last 3 observations would be considered together as a sequence. 

The key advantage of this approach is that it is easy to interpret the detectors in terms of domain knowledge. This is because at the phenotypic level they can be interpreted as conditional rules specifying intervals on the three network traffic statistics. By contrast, with the binary string representation used in LISYS there is no corresponding phenotype, and the $r$-contiguous bits matching rule does not have an intuitive interpretation at the domain level.

However, we argue that there is still a problem of scalability up to a real-world IDS with this approach. This is because only a small number of aggregate network traffic statistics have so far been considered. While such information is undoubtedly useful, it could not be used in isolation in a real system. For example, it is surely important to know discrete information such as the service ports that are being accessed. However, previous research has considered only real-valued network features; the problem of how to use a similar genotype / phenotype distinction and genetic detector generation algorithm with discrete fields has not been previously addressed. In particular, the previously unanswered questions were how to compute detector fitness with discrete fields, and how to define the similarity between two detectors, in order to produce a detector set that is spread out in antigen space. These issues are addressed in the following section of this paper, where a novel detector representation scheme and generation algorithm is presented. 

\subsection{Negative Selection Using a Genetic Algorithm}

\subsubsection{Detector Representation}
\label{secRepresentation}

The antigens in our system are connection vectors, that is, vectors storing properties of an incoming network connection. These include features such as the service port that the connection is made to and the duration of the connection. The detectors in our system then specify conditions on these features; a detector matches a connection vector if all of the conditions specified by the detector are met by the connection vector. The conditions can take one of two forms: they can either specify a single value, e.g. that the connection must be made using the UDP protocol, or they can specify an interval, e.g. that the number of connections to the same host in the last 2 seconds must be between 5 and 10. 

Significantly, a detector does not have to specify conditions on all of the features;  any of the fields can be left unspecified. Leaving fields unspecified in this way allows a detector to cover a larger area of the antigen space, potentially allowing more attacks to be detected. By contrast, in previous works such as \cite{Gonzalez:2002:c,Gonzalez:2002:j}, every field had to be specified by every detector. While this can make detector generation easier, by making it less likely that a new detector will cover any part of the self space, it has the drawback that a larger detector set will be required to cover the same amount of the non-self space. 

The genotype of a detector is then an ordered list of these conditions. For a condition specifying a single value, one genotypic field corresponds to one phenotypic field. For a condition specifying an interval, there are genotypic fields for the upper and lower bounds. 

Due to the large range of possible values for many of the fields, it was decided to use a clustering process. This was performed on all fields apart from port number by assigning the possible field values into clusters, such that each cluster contained approximately the same range of values. The port number field was handled as a special case by performing clustering manually using domain knowledge. Specifically, ports were grouped into functional categories, as shown in Table~\ref{tabPortCats}. The motivation behind this was to cluster related ports together. This could not be done by the process used for the other fields, since functionally related ports are not necessarily numbered sequentially.

\begin{table}[htbf]
\begin{center}
\fontsize{8}{10pt}\selectfont
\begin{tabular}{|l|l|}
\hline
\textbf{Category} & \textbf{Description}  \\
\hline 
1 & Remote shell   \\ \hline
 
2 & FTP \\ \hline
 
3 & HTTP \\ \hline
  
4 & Mail \\ \hline
  
5 & SQL \\ \hline
  
6 & Several ports known to be unsafe\\ \hline

7 & Network diagnostics \\ \hline
8 & 0 - 49151 (excluding those above): System / Registered ports \\ \hline

9 & 49152 - 65535: Dynamic / user-defined ports \\ \hline

\end{tabular}
\caption{Port categories used by our system.}
\label{tabPortCats}
\end{center}
\end{table}

There are two advantages to clustering field values. The first is a reduction in the size of the search space during detector generation. The second advantage follows from the fact that since a cluster represents many values, a detector specifying a cluster instead of a single value will necessarily cover a larger area of antigen space. The disadvantage of any clustering is the loss of the ability to refer to the clustered values independently. However, in the scenario dealt with in this paper, this was not deemed to be a significant problem. For example, whether 132 or 133 packets were received over a connection is unlikely to be significant in determining whether a connection is normal or anomalous.   

\subsubsection{Detector-antigen matching rule}

A detector matches a connection vector when the components of the connection vector match all of the \emph{specified} fields of the detector; any fields left unspecified by a detector are ignored. For non-interval fields, such as the port category, matching is defined simply as equality between the two values. For interval features, such as connection duration, a match occurs when the relevant component of the connection vector lies within the (inclusive) bounds specified by the corresponding lower and upper bound fields of the detector. As previously discussed, this definition of detector-antigen matching should be contrasted with the r-contiguous bits matching rule that is traditionally used in negative selection algorithms.

\subsubsection{Detector Generation} 

This section describes how a set of detectors is produced using a novel generation algorithm inspired by \cite{Gonzalez:2002:c}, but which is extended to handle discrete fields.

Detector generation is a multi-modal search problem, since individual detectors must cover different parts of the non-self space. Our algorithm follows \cite{Gonzalez:2002:c} in the use of a (steady-state) genetic algorithm with deterministic crowding \cite{Mahfoud:1992:a} for this purpose. Of particular importance is the fact that only examples of normal connections are used during detector generation, thereby avoiding biasing the generated detectors towards only recognising variants of known attacks.

A detector generation algorithm should optimise the following two objectives, computed as discrete analogues of the way that they are presented in \cite{Gonzalez:2002:c,Gonzalez:2002:j}:

\begin{enumerate}
\item $obj^1$ = \textbf{Maximise} the \textit{generality} of the detector. Generality is defined as the sum of the (normalised) ranges specified by each interval field plus the number of unspecified non-interval fields, where the result is normalised to lie between 0 and 1.
\item $obj^2$ = \textbf{Minimise} the number of self samples in the training data matched by the detector. This is negative selection.
\end{enumerate}

The purpose of $obj^1$ is to ensure that the detector set covers as large an area of the non-self space as possible, in order to be able to detect as many attacks as possible. Conversely, the purpose of $obj^2$ is to minimise coverage of the self space, in order to minimise the number of false positives. A tension therefore exists between these objectives, since the first favours detectors that cover a large area of antigen space in order to increase the attack detection rate, while the second favours those that cover a small area in order to reduce the false positive rate. Specifying the relative importance of each of these objectives allows the trade-off between the attack detection and false positive rates to be controlled. 
 
One approach for dealing with multiple objectives is to weight each objective, and then sum the weighted objective values to yield the overall fitness \cite{Bentley:1997:a}.

However, our two objectives yield values on different scales, making it inappropriate to weight them directly \cite{Bentley:1997:a}. We therefore use the Sum of Weighted Ratios method proposed in \cite{Bentley:1997:a}, which computes overall fitness from a number of different objectives as shown in Equation~\ref{eqnFitnessRatio}, where $i$ is an index over all individuals in the population, and $j$ an index over objectives. 

\begin{equation}
fr_i^j = \dfrac{(obj_i^j - min(obj^j))} {(max(obj^j) - min(obj^j))}   
\label{eqnFitnessRatio}
\end{equation}

The fitness of a detector, $i$, is then as given in Equation~\ref{eqnOverallFitness}, where $w_1$ and $w_2$ are the objective weights, and must sum to 1. The negative sign between the objectives is because $obj^2$ is a penalty for covering the self space, as defined by the examples of normal connections in the training data. 

\begin{equation}
fitness_i = w_1 * fr_i^1 - w_2*fr_i^2
\label{eqnOverallFitness}
\end{equation}

A property of the Sum of Weighted Ratios technique that makes it particularly appropriate for detector generation is that it produces best compromise solutions along a narrow range of the Pareto Front [5]. This property is useful for our application, since detectors that greatly achieve one objective to the detriment of the other, e.g. by having a large generality but consequently also matching many self samples, would not be useful. Rather, detectors are required that achieve a reasonable compromise between both objectives, i.e. that cover as large an area of antigen space as possible whilst matching few self samples in the training data. Consequently, the availability of a large range of candidate detectors spread out right across the Pareto Front is not required.

The population of detectors is initialised at the start of the algorithm as follows. Each possible phenotypic field (condition in the IF-THEN rule) is left unspecified with a probability of 0.5, represented on the detector genotype by assigning to the corresponding gene the value of -1. Leaving some fields / conditions unspecified in this way increases the generality of the detector as defined in $obj^1$. The initial values of the corresponding genes for each remaining \emph{specified} field are then chosen randomly from the list of values allowed for that gene, i.e. from the set of discrete cluster identifiers for the field.

Uniform crossover is used at each iteration of the steady-state genetic algorithm to produce a single child from two randomly chosen parents. Each gene of the resulting child is then mutated with a small probability. This mutation is performed by replacing the value of the gene with a randomly chosen value from the list of those allowed for that gene. Alternatively, the value of the gene is randomly set to -1, which means that the corresponding field is left undefined at the phenotypic level. Under the deterministic crowding scheme, the child replaces the parent that it is most similar to if it is fitter than that parent. 

Similarity between two detectors in the population is defined at the phenotypic level as follows. A similarity score is computed for each corresponding field in the two detectors. For non-interval fields, a score of 1 is given if the fields store the same value, otherwise the score is 0. For interval fields, the score is the degree of overlap between the corresponding intervals, normalised to lie between 0 and 1. The sum of the scores from each field then yields the overall similarity between the two detectors. 

At the end of the generation algorithm, the following process is executed in order to remove any detectors that still match one or more normal connections in the training set. Each detector is compared with every normal connection in the training set. If a detector matches any such connection, then that detector is discarded. Consequently, the final detector set will only contain detectors that do not match any normal training examples, thereby reducing the false positive rate.

\section{Deriving Higher-level Information About Anomalies Using a Self Organising Map}

\subsection{Creating a SOM of attack connection vectors}

The SOM is a neural network trained with a competitive learning rule in an unsupervised manner. A competitive learning rule means that the neurons compete to respond to a stimulus, such as a connection vector (recall that a connection vector describes properties of a network connection, such as the destination port and number of packets sent). The neuron that is most excited by the stimulus, i.e. whose weight vector is most similar to the connection vector, wins the competition. The winning neuron earns the right to respond to that stimulus in future, and the learning rule adjusts its weight vector so that its response to that stimulus in future will be enhanced, i.e. by moving the weight vector closer to the connection vector. This means that the next time that same connection vector is presented, the neuron that won the competition for that same vector last time will be more excited by it. 

Unsupervised learning means that the training examples are not labelled with target outputs. In our case, this means that the unsupervised learner is simply presented with connection vectors that were recorded during attacks; the learner is not told which type of attack those connection vectors were generated by. The task of the unsupervised learner is therefore to discover hidden structure or patterns in the training data. For this application, the aim is to discover clusters of similar attacks in the training data. Attacks belonging to the same cluster will then share some common properties, i.e. higher-level features.   

The training set consists of connection vectors that occurred during example attacks. It must be stressed that no examples of normal connections are included in the training data, since the decision as to whether a connection is normal or not is handled separately by the anomaly detectors described in Section~\ref{secAnomalyDetection}; the SOM only has to process connections that have already been flagged as anomalous.

At runtime, the SOM projects already identified anomalous connection vectors onto an output layer of neurons. The neurons in this layer posses a spatial topology, such as a square or rectangle, although other shapes are possible. This layer of neurons is referred to as the output grid. Each neuron has a weight vector of the same dimensionality as the connection vectors. The winning neuron for a connection vector is the neuron whose weight vector is closest to the connection vector. In this paper, a Euclidean distance metric between vectors is used. The connection vector is then projected onto the winning neuron. 

During training, the SOM learns to project connection vectors that are close together (in terms of Euclidean distance) onto neurons that are close to each in the output grid. In this way, the SOM learns relationships between the connection vectors, expressing them as spatial relationships in the output grid. The training algorithm also ensures that the weight vectors of the neurons are a good representation of the connection vectors in the training data. This is achieved by aiming for a low mean quantisation error, where the quantisation error is the distance between a connection vector and the winning neuron's weight vector. The mean quantisation error is the average of this over all connection vectors in the training set. 

The training algorithm for the SOM is as follows (adapted from the presentation given in \cite{Haykin:1999:a}):

\begin{enumerate}
	\item Initialise the weight vectors, $\textbf{w}_j(0)$, where $j$ is an index denoting the neuron number and runs from $1,2,\ldots ,l$, where $l$ is the total number of neurons in the output grid. The number in $()$ afterwards is used to denote the time-step. The weight vectors can be initialised by setting each component of each weight vector to a small random number (say between -0.1 and 0.1).
	\item Draw a connection vector, $\textbf{x}(n)$, from the training data without replacement.
	\item Find the winning neuron, $i(\textbf{x}(n))$, for the connection vector presented at this time-step. This is the neuron whose weight vector is closest to $\textbf{x}(n)$ in a Euclidean sense, calculated as follows: \[
i(\textbf{x}(n)) = \arg \mathop {\min }\limits_j \left\| {\textbf{x}(n) - \textbf{w}_j(n) } \right\|,j = 1,2,\ldots ,l
\]
	\item Adjust the weight vectors of all neurons, as follows: 
\[
\textbf{w}_j (n + 1) = \textbf{w}_j (n) + \eta (e)h_{j,i(\textbf{x})} (e)(\textbf{x}(n) - \textbf{w}_j (n)),
\]where $\eta (e)$ is the learning rate at epoch number $e$, and $h_{j,i(\textbf{x})}$ is the neighbourhood function defined in Equation~\ref{eqnNeighbour1} below, centred on $i(\textbf{x})$. Both the neighbourhood function width and learning rate vary with time, as stated in Equations \ref{eqnNeighbourhoodDecay} and \ref{eqnLR} below, respectively. Note that $n$ denotes the time-step that is incremented with the presentation of each training example, whereas $e$ denotes the epoch number. $n$ is therefore incremented after each training example has been presented, whereas $e$ is incremented after all training examples have been presented once.

	\item Repeat from step 2 until all training examples have been presented. This constitutes one epoch.
	\item Repeat from step 2 until the desired number of epochs is reached.
	
\end{enumerate}

The neighbourhood function, $h_{j,i(\textbf{x})}$, follows a Gaussian probability distribution as defined by Equations \ref{eqnNeighbour1} and \ref{eqnNeighbour2}, where $\textbf{r}_j$ denotes a vector storing the spatial position of neuron $j$ on the discrete output grid, and $\sigma ^2 (e)$ denotes the variance of the Gaussian probability distribution at epoch number $e$.
\begin{equation}
h_{j,i(\textbf{x})} (e) = \exp \left( { - \frac{{d_{j,i}^2 }}
{{2\sigma ^2 (e)}}} \right)
\label{eqnNeighbour1}
\end{equation}

\begin{equation}
d_{j,i}^2  = \left\| {\textbf{r}_j  - \textbf{r}_i } \right\|^2
\label{eqnNeighbour2} 
\end{equation}

It is important to note that the Euclidean distance in this neighbourhood equation is between the neuron positions on the output grid, and not between the corresponding weight vectors. This is necessary to ensure that neighbouring neurons in the output grid have similar weight vectors, thereby allowing a topological mapping from connection vector space to the output grid to take place. It is this concept of a neighbourhood function defined over the output grid of neurons that distinguishes this algorithm from traditional vector quantisation.

The variance of the Gaussian probability distribution can be viewed as the neighbourhood width. It therefore defines the extent to which the winning neuron's neighbours participate in the learning process by also updating their weights. In order to ensure convergence, i.e. that given enough epochs the training algorithm will eventually lead to a state where no further weight adjustments are made, it is necessary to reduce this width with increasing time. We use the exponential decay function cited in \cite{Haykin:1999:a}, presented here in Equation~\ref{eqnNeighbourhoodDecay} below. 
\begin{equation}
\sigma (e) = \sigma _0 \exp \left( { - \frac{e}
{{\tau _1 }}} \right)
\label{eqnNeighbourhoodDecay}
\end{equation}

As before, $e$ denotes the epoch number; $\sigma _0$ denotes the initial width / variance. For a square output grid of neurons, we have found it suitable to initialise the width to the length of one side of the square, e.g. in a 10-by-10 grid it would be initialised to 10. $\tau _1$ is a time constant, which following advice in \cite{Haykin:1999:a} we define as in Equation~\ref{eqnTau1}. 
\begin{equation}
\tau _1  = \frac{{1000}}
{{\log (\sigma _0 )}}
\label{eqnTau1}
\end{equation}.

In addition, to ensure convergence the learning rate, $\eta$, should also decrease with time. We again use the exponential decay suggested in \cite{Haykin:1999:a}, presented in Equation~\ref{eqnLR}, where $\eta _0 = 0.1$ is the initial learning rate, and $\tau _2 = 1000$. 
\begin{equation}
\eta (e) = \eta _0 \exp \left( { - \frac{e}
{{\tau _2 }}} \right)
\label{eqnLR}
\end{equation}

\subsubsection{Extracting Cluster Information through Labelling the SOM}
\label{secLabelSOM}

After this process of unsupervised training, it is possible to use information about the attack types of connections in the training data in order to label the neurons on the SOM output grid, i.e. label the cluster centres with higher-level information. In this paper, the attacks in the training data are grouped into broad categories; each neuron is then labelled as representing one of these categories. Specifically, the 4 broad classes of attack type defined by MIT Lincoln Labs \cite{mit:1999:a} are used, as stated below:

\begin{itemize}
	\item \emph{Denial-of-Service (DoS)}: These are attacks designed to make some service accessible through the network unavailable to legitimate users. For example, a successful DoS attack against a Web server may make the pages stored on that server unavailable for a period of time.
	\item \emph{Probe}: A Probe is a reconnaissance attack designed to uncover information about the network, which can be exploited by another attack. For example, a port scan connects to many different ports on a machine in order to determine which services that machine is running. An attacker could then look up known vulnerabilities in these services which he could exploit in a follow up attack.
	\item \emph{Remote-to-Local (R2L)}: This is where an attacker with no privileges to access a private network attempts to gain access to that network from outside, e.g. over the internet. An example would be a dictionary attack to try and guess the password of a legitimate user.
	\item \emph{User-to-Root (U2R)}: Here, the attacker has a legitimate user account on the target network. However, the attack is designed to escalate his privileges so that he can perform unauthorised actions on the network, e.g. delete another user's files.
\end{itemize}

It should be stressed that these 4 attack classes are broad, with each one encompassing many different named attacks. However, all of the attacks within a class have the same goal, and are also likely to use similar means to achieve that goal. Therefore, knowing which of these classes a (possibly new) attack belongs to tells the human security officer the purpose of the attack and the actions that it may involve. 

The neurons of the trained SOM are labelled with these attack classes using the algorithm shown below:

\begin{enumerate}
	\item Draw an attack connection vector from the training set, without replacement.
	\item Find the winning neuron for that connection vector. 
	\item Add 1 to the count of the number of connection vectors of that class projected onto that neuron.
	\item Repeat from 1 until all connection vectors in the training set have been presented.
	\item Label each neuron with the most frequently occurring class of connection vectors that was projected onto it.
\end{enumerate}

After labelling, when a new connection vector is projected onto the SOM then it is assigned to the attack class of the corresponding winning neuron.

\subsubsection{Improving Attack Classification Performance Using Learning Vector Quantisation}
\label{secLVQ}

Learning Vector Quantisation (LVQ) \cite{Kohonen:1990:b} adjusts the weight vectors of the SOM after the initial unsupervised training. Initially, the neurons in the SOM are labelled with the attack classes they represent, as previously described in Section~\ref{secLabelSOM}. After labelling, the weight vector of the winning neuron for each training example is adjusted as follows. If the winning neuron has the same attack class label as the connection vector, then the weight vector is moved towards that connection vector. Alternatively, if the winning neuron and connection vector have different class labels then the weight vector is moved away from the connection vector that has just been misclassified. This process repeats, looping through the whole training set several times. A full presentation of the algorithm is given below, adapted from the presentation given in \cite{Haykin:1999:a}:

Present each labelled connection vector in the training set, one at a time. Let $\textbf{x}(n)$ denote the connection vector presented at time-step $n$, and let $e$ denote the epoch number. At each time-step, make one of the following adjustments to the weight vector $\textbf{w}_{i(\textbf{x}(n))}$ of the winning neuron $i(\textbf{x}(n))$:
\begin{itemize}
	\item If the class label of the winning neuron is the same as the class label of $\textbf{x}(n)$ then:
\[	 
\textbf{w}_{i(\textbf{x}(n))} (n + 1) = \textbf{w}_{i(\textbf{x}(n))} (n) + \alpha _e \left[ {\textbf{x}(n) - \textbf{w}_{i(\textbf{x}(n))} (n)} \right]
\]
\item Alternatively, if the class label of the winning neuron is different to the class label of $\textbf{x}(n)$ then:
\[
\textbf{w}_{i(\textbf{x}(n))} (n + 1) = \textbf{w}_{i(\textbf{x}(n))} (n) - \alpha _e \left[ {\textbf{x}(n) - \textbf{w}_{i(\textbf{x}(n))} (n)} \right]
\]
 
\end{itemize}

The presentation of all training examples constitutes one epoch. The procedure is repeated for a specified number of epochs. $\alpha_e$ denotes the learning rate at epoch number $e$, which must be greater than 0 and less than 1. It should decrease with increasing epochs, in order to ensure convergence. 

The aim of LVQ is to reduce the number of misclassifications made on the connection vectors in the training set. It should be noted that it is a supervised learning procedure, as the class labels of the examples in the training set must be known. However, prior to LVQ, the weights of the SOM are trained in the normal unsupervised way. As far as we are aware, our system is the first IDS with a SOM component that makes use of LVQ. 

\subsection{Motivation for using a SOM for attack type classification}

A more conventional classifier, such as a Multi-Layer Perceptron, could have been trained to classify connection vectors into one of the 4 predefined broad attack classes. However, a SOM approach can be regarded as more flexible for several reasons. The first reason is that the operation of the SOM is very transparent compared to that of a Multi-Layer Perceptron. This is because a Multi-Layer Perceptron operates as a black box, making it very difficult to see why an anomalous connection vector is classified as being of a particular attack type. With a SOM, however, the results of the training process can easily be visualised by plotting the output grid of neurons and labelling them with attack classes. A new anomalous connection can then be projected onto this grid graphically, revealing which neurons and hence classes it is closest to.  

Another advantage of the SOM is that it is not required that the attack connection vectors in the training data be grouped into predefined classes. This is because the initial training of the SOM is unsupervised. The SOM will therefore group similar attack connection vectors together without reference to class labels. It was decided to use 4 broad attack class labels throughout this paper for reasons of convenience during testing, and in order to facilitate a comparison of results with other works. However, a real implementation in an IDS may benefit from providing more fine grained information. This could be in the form of details of a sample of specific named attacks that the new anomalous connection was most similar to. A SOM approach could easily accommodate this, by recording a sample of which attacks in the training data were projected onto which neuron. Then, when a new anomalous connection was projected onto a neuron, details about other attacks that have been projected onto the same neuron could be presented to the human security officer. By looking at detailed information about those other attacks, it may be possible to infer some properties about the new attack. A Multi-Layer Perceptron, however, would be unable to do this. It should also be noted that the neurons of the SOM would only have to be labelled with a sample of the details of particular attacks; a list of every attack in the training data would not have to be permanently stored, saving considerable space over a simple nearest-neighbour approach.

\section{Results}

This section presents experimental results detailing the performance of our hybrid system on the ``corrected'' 10\% version of the popular KDD 1999 Cup dataset \cite{kdd:1999:a}. This dataset consists of a set of preprocessed network connection records, each containing 41 features. The fact that the raw network traffic comprising this dataset has already been processed in order to compute statistics such as the number of failed login attempts makes this dataset a popular choice amongst IDS researchers, since it avoids the need for extensive preprocessing. For this reason, we do not address the preprocessing of raw network traffic data in this paper.

\subsection{Choice of network features and description of detectors}
\label{secDetDetails}

\subsection{Dataset description}  

The ``corrected'' KDD 1999 Cup dataset is supplied pre-partitioned into a training set consisting of (approximately):

\begin{itemize}
	\item 56000 normal connections,
	\item 8000 DoS,
	\item 3300 Probe,
	\item 29 U2R,
	\item 110 R2L,
\end{itemize}

and a testing set containing (approximately):

\begin{itemize}
	\item 19100 normal connections,
	\item 73100 DoS,
	\item 2300 Probe,
	\item 19 U2R,
	\item 1000 R2L.
\end{itemize} 

In order to facilitate comparison with existing systems, it was decided that this particular partitioning into training and testing sets should not be modified for the experimental analysis in this paper.   

A list of the network features used from the dataset along with a specification of the detector genotype are provided in Appendix~\ref{app1}.

\subsection{Anomaly Detection Performance}
\label{secAISRes}
Results on the testing set for the artificial immune system anomaly detection component are shown in Table~\ref{tabAIS}, averaged over 50 runs. This experiment focussed on varying the objective weights; the authors' previous work \cite{Powers:2006:b} has explored the effect of varying the population size of the detector generation algorithm.

The following parameter settings were held constant throughout:

\begin{itemize}
	\item population size of 1600,
	\item steady-state genetic algorithm executed for 50000 iterations,
	\item crossover rate = 1.0,
	\item mutation rate = $1/L$, where $L$ is the number of fields in the detector genotype.
\end{itemize}
 
\begin{table}[htbf]
	\centering
		\begin{tabular} {|p{1.3cm}|p{1.3cm}|p{1.5cm}|p{1.2cm}|p{1.2cm}|p{1.2cm}|p{1.2cm}|} \hline
\textbf{$obj^1$ weight} & \textbf{$obj^2$ weight} & \textbf{Normal} & \textbf{DoS} & \textbf{Probe} & \textbf{U2R} & \textbf{R2L} \\ \hline 
0.1 & 0.9 & 99.7\% & 59.2\% & 39.1\% & 42.3\% & 9.8\% \\ \hline
0.2 & 0.8 & 99.6\% & 71.1\% & 41.9\% & 47.7\% & 12.3\% \\ \hline
0.3 & 0.7 & 99.6\% & 67.3\% & 54.4\% & 50.6\% & 13.1\% \\ \hline
0.4 & 0.6 & 99.6\% & 81\% & 54.3\% & 52.1\% & 10.9\% \\ \hline
0.5 & 0.5 & 99.5\% & 91\% & 63.4\% & 58\% & 12.8\% \\ \hline
0.6 & 0.4 & 99.4\% & 96.9\% & 68.5\% & 58.4\% & 12.5\% \\ \hline
0.7 & 0.3 & 99.5\% & 94.9\% & 71.8\% & 58.9\% & 12\% \\ \hline
0.8 & 0.2 & 99.5\% & 96.9\% & 67.4\% & 58\% & 11.4\% \\ \hline
0.9 & 0.1 & 99.5\% & 98.9\% & 65.4\% & 60.3\% & 11.6\% \\ \hline
\end{tabular}
\caption{Anomaly detection performance by the artificial immune system on the ``corrected'' KDD99 test set.}
	\label{tabAIS}
\end{table}

When interpreting these results, it is crucial to understand that although attack detection performance is listed by attack class, the artificial immune system component simply classifies each connection as normal or anomalous, with no regard to class division between anomalous connections. However, the results are presented broken down into attack classes, since this reveals the anomaly detection performance on different kinds of attacks.

For all of the weight settings, the resulting false positive rate (number of normal connections misclassified) was very low, never stepping above 0.6\%. This was achieved in part due to the fact that at the end of detector generation, our algorithm destroys any detectors that still match any normal connection vectors in the training data. Such a low false positive rate is highly desirable, since it avoids overwhelming a security officer with false alarms. 

The general trend in the results is that increasing the weight of $obj^1$ increases the anomalous connection detection performance. This is intuitive, since $obj^1$ is to minimise the number of conditions in the detector IF-THEN matching rule. A rule with fewer conditions will be able to match more connection vectors in total, thereby allowing it to match more attack connection vectors. The results also show that there is not a substantial difference in performance between different settings of $obj^1 > 0.5$.

\subsection{Classification of Detected Attacks}
\label{secSOMRes}

This section reports the attack classification performance of the SOM. The task here is not to distinguish between normal and anomalous; that is handled by the artificial immune system. Instead, given an anomalous connection that has already been detected, the job of the SOM is to classify the anomaly into one of the 4 attack types.

The following parameters of the SOM training algorithm were used throughout:

\begin{itemize}
	\item Square output grid topology,
	\item 2000 epochs of training,
	\item $\eta_0 = 0.1$,
	\item $\sigma_0$ = length of one side of output grid.
\end{itemize}

\subsubsection{Varying the Size of the Output Grid}

Table~\ref{tabSomSize} shows the results of varying the size of the output grid, averaged over 2 runs. It should be noted that the only stochastic element of the SOM training algorithm is the initialisation of the neurons' weight vectors. However, the training algorithm is known to be quite insensitive to the setting of this \cite{Kayacik:2003:a}. Furthermore, any suboptimal training results due to a poor initialisation would become apparent as a topological defect in the final mapping between connection vector space and the output grid of neurons, allowing such poor runs to be discarded. For these reasons, it is not essential to repeat this experiment for a large number of times, as would be the case with, say, the back-propagation training algorithm. 

\begin{table}[htbf]
	\centering
		\begin{tabular} {|p{1.4cm}|p{1.2cm}|p{1.2cm}|p{1.2cm}|p{1.2cm}|} \hline
\textbf{Output grid size} & \textbf{DoS} & \textbf{Probe} & \textbf{U2R} & \textbf{R2L} \\ \hline 
5-by-5 & 99.13\% & 96.32\% & 0\% & 99.83\% \\ \hline
7-by-7 & 99.90\% & 93.46\% & 28.95\% & 34.84\% \\ \hline
10-by-10 & 99.57\% & 99.62\% & 55.76\% & 35.89\% \\ \hline
15-by-15 & 99.91\% & 93.18\% & 21.05\% & 37.66\% \\ \hline

\end{tabular}
\caption{Attack classification performance by SOMs of varying sizes on the test set.}
	\label{tabSomSize}
\end{table}

The results show that a SOM with a small 5-by-5 output grid of neurons produced very good correct classification scores of 99.13\%, 96.32\% and 99.83\% on the DoS, Probe and R2L attack classes, respectively. The reason for the 0\% score for the U2R class was discovered by examining a plot of the labelled output grid, which revealed that no neurons were labeled as U2R. A further examination of the winning neurons for test examples of this class revealed that they were all labelled as R2L, i.e. that U2R attacks are misclassified as R2L. This is somewhat intuitive, since both of these types of attack involve a user attempting to gain unauthorised access to a system. The difference between them is that in an R2L attack the attacker is operating from outside the network, whereas in a U2R attack he is operating from within it. 

Using a larger output grid, i.e. more neurons, allows some neurons to be labelled as representing the U2R class. For example, in a 7-by-7 grid, 28.95\% of the U2R attack connections were correctly classified. When the grid size was increased to 10-by-10, the classification performance for this type of attack improved further to 55.76\%. Unfortunately, classification performance for the R2L class decreases substantially with these larger grid sizes, being close to 35\% in both cases. An analysis of the winning neurons for the test R2L connection vectors revealed that many of the vectors were being projected onto neurons labelled as U2R. This therefore leads to the conclusion that there is an inherent trade-off in the ability to classify the two attack types correctly, with the 5-by-5 SOM forcing this trade-off to be taken to the extreme. It may be possible to remove this trade-off by using additional properties of network connections to discrminate between the classes.

\begin{figure}[htbf]
	\epsfig{file=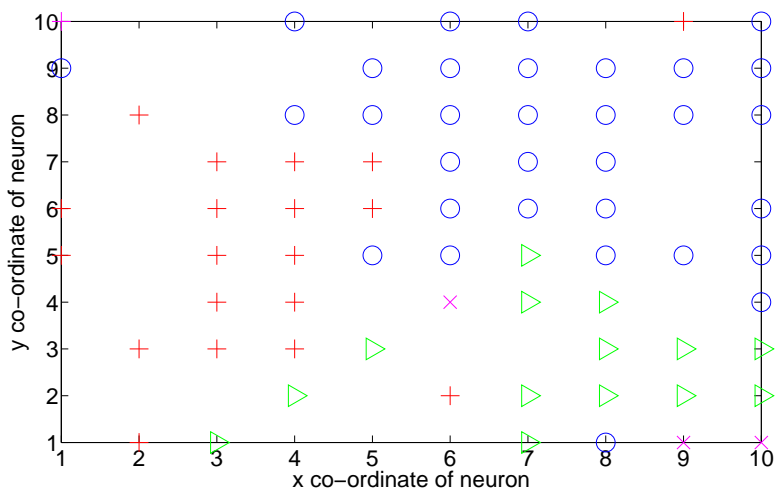,width=12cm}
% papRoc.eps: 300dpi, width=9.13cm, height=4.93cm, bb= -233   104   845   686
	\caption{The labelled output grid of neurons for a trained 10-by-10 SOM. The circles represent Probe, the +'s DoS, the triangles R2L and the x's U2R.}
	\label{10by10}
\end{figure}

From Table~\ref{tabSomSize}, the best compromise seems to be to use a grid of size 10-by-10 (for which a plot of the labelled output grid is provided in Figure~\ref{10by10}). This is because this size performs better than a 7-by-7 grid in all attack classes, except for DoS. However, the difference between their DoS scores is only 0.33\%. Although a 5-by-5 grid achieves a very high score for the R2L class, it achieves 0\% on the U2R class. It would therefore be unwise to use such a system, as it would leave the network completely vulnerable to these kinds of attack. 

\subsubsection{Testing the LVQ Algorithm}

The aim of this experiment was to determine whether or not an application of the LVQ algorithm after SOM training improves classification performance in this domain. Such a study is novel, as we are unaware of any other works that have applied LVQ in the intrusion detection domain.

Before applying LVQ, the SOM is trained in the usual manner. The nodes are then labelled, and the LVQ algorithm described in Section~\ref{secLVQ} is applied. 

Table~\ref{lvqTab} shows the effect of applying the LVQ algorithm to a SOM with an output grid of size 10-by-10. Recall that the results in Table~\ref{tabSomSize} revealed this size grid to yield a good compromise between correctly classifying the different types of attacks. The LVQ algorithm is applied for 10 epochs, with the initial learning rate varied as specified in the table. The learning rate is reduced by a half at the end of each epoch.

\begin{table}[htbf]
	\centering
		\begin{tabular} {|p{1.6cm}|p{1.2cm}|p{1.2cm}|p{1.2cm}|p{1.2cm}|} \hline
\textbf{Initial learning rate $\eta_0$} & \textbf{DoS} & \textbf{Probe} & \textbf{U2R} & \textbf{R2L} \\ \hline 
0 (no LVQ) & 99.57\% & 99.62\% & 55.76\% & 35.89\% \\ \hline
0.2 & 99.92\% & 94.46\% & 59.21\% & 41.49\% \\ \hline
0.4 & 99.90\% & 99.47\% & 28.95\% & 44.13\% \\ \hline
0.6 & 99.90\% & 99.39\% & 19.74\% & 99.77\% \\ \hline
0.8 & 99.89\% & 99.49\% & 2.63\% & 99.79\% \\ \hline

\end{tabular}
\caption{Attack classification performance for a 10-by-10 SOM after an application of the LVQ algorithm.}
	\label{lvqTab}
\end{table}

It can be seen from the results that adjusting the initial learning rate for the LVQ algorithm allows the trade-off between classifying the U2R and R2L attacks to be set. Setting a higher initial learning rate favours correct classification of R2L attacks, while a lower initial learning rate favours correct classification of the U2R attacks. The reason for this is that there are many more examples of R2L connections in the training set than their are U2R (110 versus 29). If there are more examples of one class then the LVQ algorithm will bias the improvement in classification performance towards that class. This follows from the fact that the standard LVQ algorithm treats each training example equally, and adjusts the SOMs weights to improve classification performance on each such example in turn. Therefore, the sum of the weight changes made by LVQ over all training examples will be biased towards the most frequently occurring class.

A good compromise between detecting the different types of attacks was obtained with an initial learning rate of 0.2. Compared to not using vector quantisation, this gives a 3.45\% improvement in classification performance on U2R attacks, and a 5.6\% improvement on R2L attacks. For DoS, the improvement is marginal at 0.35\%, while for Probe there is a decrease of 5.16\%. However, there is still a net improvement in classification performance across all attack types of 4.24\%. Furthermore, it can be argued that the improvement in classifying U2R and R2L attacks is more important pragmatically than the decrease in classification performance of Probe attacks, since Probe attacks were still correctly classified 94.46\% of the time. 

\subsection{Overall Performance}

Results are presented in Table~\ref{compTab} that compare the overall (detection and classification) performance of the hybrid system to that of two monolithic systems chosen from the literature. These are a recent work that uses a 3-layer hierarchy of SOMs for both attack detection and classification on the same 10\% dataset \cite{Kayacik:2003:a}, and the winning entry from the KDD 1999 Cup competition \cite{Kdd:1999:b}. The first system was chosen because it also uses a SOM, while the second was chosen as a standard benchmark reference result. 

It should be remarked that a plethora of other techniques have been applied to the KDD 1999 Cup dataset, including, for example, hybrid machine learning techniques \cite{Shon:2007:a}, genetic programming \cite{Song:2003:a}, and an anomaly detection scheme based on Principle Components Analysis \cite{Shyu:2003:a}. Likewise, various artificial immune system approaches have also been applied (of which a comprehensive review can be found in \cite{Kim:2007:a}), including a recent use of the negative selection algorithm described in \cite{Haag:2007:a}. However, it is unfortunately not easy to directly compare the performance of our system to these approaches, since their authors each calculate performance in a different way. In particular, the traditional approach from the original KDD competition of considering performance on the 4 attack classes separately is often not used; instead, all attack types are often treated homogeneously, or they are grouped in a different manner. Consequently, it was only possible to compare the performance of our system to those where the authors report the same performance statistics, from which the most relevant have been selected.   

Following from the results presented in Sections~\ref{secAISRes} and \ref{secSOMRes}, the following parameter settings were chosen for the hybrid system:

\begin{itemize}
	\item Objective weights of 0.6 \& 0.4 (for $obj^1$ and $obj^2$, respectively) for anomaly detector generation,
	\item a SOM grid of size 10-by-10,
	\item an initial learning rate of 0.2 for LVQ.
\end{itemize}
 
The overall detection and classification performance was then computed accordingly from Tables~\ref{tabAIS} and \ref{lvqTab}, i.e. by multiplying the detection rate from Table~\ref{tabAIS} with the classification rate from Table~\ref{lvqTab}.

\begin{table}[htbf]
	\centering
		\begin{tabular} {|p{2.8cm}|p{1.2cm}|p{1.2cm}|p{1.2cm}|p{1.2cm}|p{1.2cm}|} \hline
\textbf{Approach} & \textbf{Normal} & \textbf{DoS} & \textbf{Probe} & \textbf{U2R} & \textbf{R2L} \\ \hline 
Hybrid artificial immune system and SOM & 99.4\% & 96.8\% & 64.7\% & 34.6\% & 5.2\% \\ \hline
Hierarchy of SOMs (3 layers) \cite{Kayacik:2003:a} & 95.4\% & 95.1\% & 64.3\% & 10.0\% & 9.9\% \\ \hline
Winning entry from KDD 1999 Cup \cite{Kdd:1999:b} & 99.5\% & 83.3\% & 97.1\% & 13.2\% & 8.4\% \\ \hline
\end{tabular}
\caption{Comparing the performance of our hybrid system to a monolithic hierarchy of SOMs \cite{Kayacik:2003:a} and to the winning entry from the KDD 1999 Cup competition \cite{Kdd:1999:b}.}
	\label{compTab}
\end{table}

The results show that the hybrid system is better at detecting and classifying DoS and U2R attacks than the 3 layer hierarchy of SOMs. Furthermore, a lower false positive rate is also obtained; this may be because of a divide-and-conquer advantage in separating detection from classification. This enables the anomaly detection component of the hybrid system, in the form of the artificial immune system, to concentrate on having a low false positive rate without having to learn to (sub)classify attacks. Likewise, the SOM component can be trained solely to classify different kinds of attacks, without having to also learn the difference between normal and anomalous.

When comparing results with the winning entry from the KDD 1999 Cup competition, it should be noted that both the hybrid system and the hierarchy of SOMs were tested only on the 10\% version of the full dataset. However, many researchers test their systems on this same 10\%, making it a reasonable portion of the dataset to use. Bearing this in mind, the results show that the hybrid system is better at detecting and classifying DoS (96.8\% versus 83.3\%) and U2R (34.6\% versus 13.2\%) attacks than the winning entry. In addition, the false positive rates of both systems are similar. 

However, the winning entry is better at detecting and classifying Probe (97.1\% versus 64.7\%) and R2L (8.4\% versus 5.2\%). Using more features in the connection vectors may help to improve the performance of the hybrid system on these attacks. Finally, the fact that the winning entry also has low detection and classification rates for U2R and R2L attacks suggests that such attacks are indeed problematic for intrusion detection systems. In addition, if classification of these two attack types is considered a trade-off, as previously discussed, then the results of the hybrid system are favourable. This is because the hybrid system is 2.6 times better at detecting U2R attacks, whereas the winning entry is only 1.6 times better at detecting R2L attacks.

\section{Conclusion}

The main contribution of this paper has been to propose a hybrid IDS that combines pure anomaly detection with the provision of higher-level information about the detected anomalies. At present, this information takes the form of the broad attack type. However, other sorts of information are possible, such as a list of specific example attacks that the anomaly is most similar to. By contrast, other works that propose an anomaly detection based IDS typically only provide low-level output. This low-level output is usually a binary score indicating whether a connection is anomalous or not. However, such low-level information leaves a security officer with much work to do to find out the possible actions and consequences of an attack. 

The key feature and novelty of the system presented in this paper is the use of separate components for anomaly detection and attack classification. Detection is about recognising that a connection is anomalous, while classification is about determining the broad attack type of the connection. The advantage of using separate components is that in the first stage, the system is able to perform pure anomaly detection. In other words, examples of specific attacks are not required for attack detection, thereby avoiding biasing the system towards only detecting variants of known attacks. This is important, given the rate at which new attacks are developed. This approach should be contrasted with the misuse based approach that looks for signatures of known attacks, since such systems cannot detect attacks for which a signature is not present. 

Comparing the performance of the system described in this paper to a sample of other works on the KDD 1999 Cup dataset has shown favourable false positive and attack classification results. It would be valuable in future to consider the attack detection rate on novel attacks only, since the use of anomaly detection may provide a further advantage against such attacks. Finally, regarding the nature-inspired techniques used in our system, the system presented here is the first to combine an artificial immune system with another technique at runtime as part of a hybrid IDS.

\section*{Acknowledgements}

This research was conducted while both authors were members of the School of Computer Science, University of Birmingham. The first author was funded by the MSc. Natural Computation programme at the University of Birmingham.

\appendix
\section{Detector Genotype for the KDD 1999 Cup Dataset}
\label{app1}

The 18 features shown in Table~\ref{tabFeatures} were selected for use from the KDD 1999 Cup dataset. A clustering process was performed on the discrete and real-valued fields in order to reduce the size of the detector search space and increase detector generality, as previously discussed in Section~\ref{secRepresentation}. This clustering was carried out using the standard equal-frequency binning algorithm, which divides the values observed in the training data into a number of bins, such that approximately the same number of training records are placed in each bin. The number of bins used for each feature is shown in the third column of Table~\ref{tabFeatures}. 

\begin{table}[htbf]
	\centering
		\begin{tabular} {|c|c|c|} \hline
\textbf{Feature} & \textbf{Datatype} & \textbf{Number of bins} \\ \hline 
Connection duration & integer & 8\\ \hline
Protocol type & categorical & N/A\\ \hline
Port category & integer & 9\\ \hline
Number of urgent packets & integer & 3\\ \hline
Number of ``hot'' indicators in packet contents & integer & 3\\ \hline
Number of failed login attempts & integer & 3\\ \hline
Whether the user is logged in successfully & binary & N/A\\ \hline
Whether a root shell was obtained & binary & N/A\\ \hline
Whether the command ``su root'' was attempted & binary & N/A\\ \hline
Number of file creation operations & integer & 4\\ \hline
Number of open shell prompts & integer & 3\\ \hline
Whether the login username belongs to the ``hot list'' & binary & N/A\\ \hline
Whether the login is to a guest account & binary & N/A \\ \hline
Number of connections to same host in past 2 seconds & integer & 10\\ \hline
\% of connections to same port in past 2 seconds & real & 3\\ \hline
\% of connections to different ports in past 2 seconds & real & 3\\ \hline
\% of connections to different hosts in past 2 seconds & real & 3\\ \hline

\end{tabular}
\caption{Network features used from KDD 1999 Cup Dataset.}
	\label{tabFeatures}
\end{table}

Given this choice of network features, a detector genotype is then of the following form:

\begin{enumerate}
\item Lower bound on the connection duration,
\item upper bound on the connection duration,
\item protocol type,
\item port category,
\item connection from/to same host/port?,
\item lower bound on number of urgent packets,
\item upper bound on number of urgent packets,
\item lower bound on number of ``hot'' indicators,
\item upper bound on number of ``hot'' indicators,
\item lower bound on number of failed login attempts,
\item upper bound on number of failed login attempts,
\item user logged in?,
\item root shell obtained?,
\item 'su root' attempted?,
\item lower bound on number of file creation operations,
\item upper bound on number of file creation operations,
\item lower bound on number of shell prompts open,
\item upper bound on number of shell prompts open,
\item login user name in ``hot list''?,
\item guest login?,
\item lower bound on number of connections to same host in past 2 seconds,
\item upper bound on number of connections to same host in past 2 seconds,
\item lower bound on \% of connections to same port in past 2 seconds,
\item upper bound on \% of connections to same port in past 2 seconds,
\item lower bound on \% of connections to different ports in past 2 seconds,
\item upper bound on \% of connections to different ports in past 2 seconds,
\item lower bound on \% of connections to different hosts in past 2 seconds,
\item upper bound on \% of connections to different hosts in past 2 seconds.
\end{enumerate}

It should be noted that any of these fields can be left unspecified, denoted by a value of -1.

\bibliographystyle{elsart-num-sort}

\begin{thebibliography}{10}
\expandafter\ifx\csname url\endcsname\relax
  \def\url#1{\texttt{#1}}\fi
\expandafter\ifx\csname urlprefix\endcsname\relax\def\urlprefix{URL }\fi

\bibitem{Balthrop:2002:a}
J.~Balthrop, S.~Forrest, M.~Glickman, Revisiting {LISYS}: Parameters and normal
  behavior, in: Proceedings of the 2002 Congress on Evolutionary Computation
  (CEC'02), vol.~2, IEEE Press, 2002.

\bibitem{Bass:1999:a}
T.~Bass, Multisensor data fusion for next generation distributed intrusion
  detection systems, in: Proceedings of the 1999 IRIS National Symposium on
  Sensor and Data Fusion, 1999.

\bibitem{Bass:2000:a}
T.~Bass, Intrusion detection systems and multisensor data fusion,
  Communications of the ACM 43~(4) (2000) 99--105.

\bibitem{Bentley:1997:a}
P.~J. Bentley, J.~P. Wakefield, Finding acceptable solutions in the
  {Pareto}-optimal range using multiobjective genetic algorithms, in: P.~K.
  Chawdhry, R.~Roy, R.~K. Pant (eds.), Soft Computing in Engineering Design and
  Manufacturing, Springer, 1997.

\bibitem{Cannady:1998:a}
J.~Cannady, Artificial neural networks for misuse detection, in: Proceedings of
  the 1998 National Information Systems Security Conference (NISSC'98), 1998.

\bibitem{Snort:2006:a}
B.~Caswell, M.~Roesch, {The SNORT network intrusion detection system},
  http://www.snort.org (2006).

\bibitem{deCastro:2002:a}
L.~N. de~Castro, J.~Timmis, {Artificial Immune Systems}: A New Computational
  Intelligence Approach, Springer, 2002.

\bibitem{DeLooze:2004:a}
L.~L. DeLooze, Classification of computer attacks using a self-organizing map,
  in: Proceedings of the Fifth Annual IEEE SMC Information Assurance Workshop,
  IEEE Press, 2004.

\bibitem{DeLooze:2006:a}
L.~L. DeLooze, Attack characterization and intrusion detection using an
  ensemble of self-organizing maps, in: Proceedings of the 2006 IEEE
  Information Assurance Workshop, IEEE Press, 2006.

\bibitem{Denning:1987:a}
D.~Denning, An intrusion detection model, IEEE Transactions on Software
  Engineering 13~(2) (1987) 222--232.

\bibitem{Depren:2005:a}
O.~Depren, M.~Topallar, E.~Anarim, M.~K. Ciliz, An intelligent intrusion
  detection system {(IDS)} for anomaly and misuse detection in computer
  networks, Expert Systems with Applications 29~(4) (2005) 713--722.

\bibitem{Forrest:1996:a}
S.~Forrest, S.~A. Hofmeyr, A.~Somayaji, T.~A. Longstaff, A sense of self for
  {Unix} processes, in: Proceedings of the 1996 IEEE Symposium on Security and
  Privacy, IEEE Press, 1996.

\bibitem{Forrest:1994:a}
S.~Forrest, A.~Perelson, L.~Allen, R.~Cherukuri, Self-nonself discrimination in
  a computer, in: Proceedings of the 1994 IEEE Symposium on Research in
  Security and Privacy, IEEE Press, 1994.

\bibitem{Gonzalez:2002:j}
F.~Gonz\'{a}lez, D.~Dasgupta, An immunity-based technique to characterize
  intrusions in computer networks, IEEE Transactions on Evolutionary
  Computation 6~(3) (2002) 281--291.

\bibitem{Gonzalez:2002:c}
F.~Gonz\'{a}lez, D.~Dasgupta, An imunogenetic technique to detect anomalies in
  network traffic, in: Proceedings of the Genetic and Evolutionary Computation
  Conference (GECCO), Morgan Kaufmann, 2002.

\bibitem{Gonzalez:2003:b}
F.~Gonz\'{a}lez, D.~Dasgupta, Anomaly detection using real-valued negative
  selection, Genetic Programming and Evolvable Machines 4~(4) (2003) 383--403.

\bibitem{Haag:2007:a}
C.~R. Haag, G.~B. Lamont, P.~D. Williams, G.~L. Peterson, An artificial immune
  system-inspired multiobjective evolutionary algorithm with application to the
  detection of distributed computer network intrusions, in: Proceedings of the
  6th International Conference on Artificial Immune Systems (ICARIS 2007), vol.
  4628 of Lecture Notes in Computer Science, Springer, 2007.

\bibitem{Haykin:1999:a}
S.~Haykin, Neural Networks: A Comprehensive Foundation, Prentice-Hall, 1999.

\bibitem{Hofmeyr:1999:a}
S.~A. Hofmeyr, S.~Forrest, Immunity by design: an {Artificial Immune System},
  in: Proceedings of the Genetic and Evolutionary Computation Conference
  (GECCO'99), vol.~2, Morgan Kaufmann, 1999.

\bibitem{Kayacik:2003:a}
H.~G. Kayacik, A.~N. Zincir-Heywood, M.~I. Heywood, On the capability of an
  {SOM} based intrusion detection system, in: Proceedings of the 2003
  International Joint Conference on Neural Networks, vol.~3, IEEE Press, 2003.

\bibitem{Kim:2001:b}
J.~Kim, P.~J. Bentley, Evaluating negative selection in an {Artificial Immune
  System} for network intrusion detection, in: Proceedings of the 2001 Genetic
  and Evolutionary Computation Conference (GECCO'01), Morgan Kaufmann, 2001.

\bibitem{Kim:2007:a}
J.~Kim, P.~J. Bentley, U.~Aickelin, J.~Greensmith, G.~Tedesco, J.~Twycross,
  Immune system approaches to intrusion detection – a review, Natural Computing
  6~(4) (2007) 413--466.

\bibitem{Kohonen:1982:a}
T.~Kohonen, Self-organized formation of topologically correct feature maps,
  Biological Cybernetics 43 (1982) 59--69.

\bibitem{Kohonen:1990:b}
T.~Kohonen, Improved versions of learning vector quantization, in: Proceedings
  of the IEEE International Joint Conference on Neural Networks, vol.~1, IEEE
  Press, 1990.

\bibitem{Li:2004:a}
J.~Li, G.-Y. Zhang, G.-C. Gu, The research and implementation of intelligent
  intrusion detection system based on artificial neural network, in:
  Proceedings of 2004 International Conference on Machine Learning and
  Cybernetics, vol.~5, IEEE Press, 2004.

\bibitem{Mahfoud:1992:a}
S.~W. Mahfoud, Crowding and preselection revisited, in: Proceedings of the
  Second Conference on Parallel Problem Solving from Nature, North-Holland,
  1992.

\bibitem{mit:1999:a}
{MIT Lincoln Labs}, 1999 {DARPA} intrusion detection evaluation, available
  online at: http://www.ll.mit.edu/IST/ideval/.

\bibitem{Mukherjee:1994:a}
B.~Mukherjee, L.~T. Heberlein, K.~N. Levitt, Network intrusion detection, IEEE
  Network 8~(3) (1994) 26--41.

\bibitem{Mukkamala:2002:a}
S.~Mukkamala, G.~Janoski, A.~Sung, Intrusion detection using neural networks
  and support vector machines, in: Proceedings of the 2002 International Joint
  Conference on Neural Networks (IJCNN'02), vol.~2, IEEE Press, 2002.

\bibitem{Kdd:1999:b}
B.~Pfahringer, Winning entry of the kdd'99 classifier learning contest, results
  available online at: http://www.acm.org/sigs/sigkdd/kddcup/ (1999).

\bibitem{Powers:2006:b}
S.~T. Powers, J.~He, Evolving discrete-valued anomaly detectors for a network
  intrusion detection system using negative selection, in: X.~Z. Wang, R.~F. Li
  (eds.), Proceedings of the 2006 UK Workshop on Computational Intelligence
  (UKCI 2006), University of Leeds, 2006.

\bibitem{Shon:2007:a}
T.~Shon, J.~Moon, A hybrid machine learning approach to network anomaly
  detection, Information Sciences 177~(18) (2007) 3799--3821.

\bibitem{Shyu:2003:a}
M.-L. Shyu, S.-C. Chen, K.~Sarinnapakorn, L.~Chang, A novel anomaly detection
  scheme based on principal component classifier, in: Proceedings of the IEEE
  Foundations and New Directions of Data Mining Workshop, IEEE Press, 2003.

\bibitem{Song:2003:a}
D.~Song, M.~I. Heywood, A.~N. Zincir-Heywood, A linear genetic programming
  approach to intrusion detection, in: Proceedings of the 2003 Genetic and
  Evolutionary Computation Conference (GECCO 2003), Springer, 2003.

\bibitem{Cert:2006:a}
{The Software Engineering Institue at Carnegie Mellon University}, Cert/cc
  statistics 1988-2006, http://www.cert.org/stats/ (2006).

\bibitem{kdd:1999:a}
{The UCI KDD Archive}, {KDD Cup} 1999 data, available online at:
  http://kdd.ics.uci.edu//databases/kddcup99/kddcup99.html.

\end{thebibliography}

\end{document}